# MRA - Proof of Concept of a Multilingual Report Annotator Web Application

Luís Campos, Francisco Couto
LaSIGE, University of Lisbon, Portugal

**Abstract**

MRA (Multilingual Report Annotator) is a web application that translates Radiology text and annotates it with RadLex terms. Its goal is to explore the solution of translating non-English Radiology reports as a way to solve the problem of most of the Text Mining tools being developed for English. In this brief paper we explain the language barrier problem and shortly describe the application. MRA can be found at https://github.com/lasigeBioTM/MRA.

## Background

Radiology reports describe the results of radiography procedures and have the potential of being an useful source of information (1), which can bring benefits to health care systems around the world. However, these reports are usually written in free-text and thus it is hard to automatically extract information from them. Nonetheless, the fact that most reports are now digitally available makes them amenable for using Text Mining tools. Another advantage of Radiology reports is that even if written in free-text, they are usually well structured. A lot of work has been done on Text Mining of Biomedical texts, including health records (2), but although Radiology reports are usually written in the native language of the radiologist, Text Mining tools are mostly developed for English. For example, Hassanpour *et al. (3)* created an information extraction system for English reports that depends on RadLex (4), a lexicon for Radiology terminology, which is freely available in English. Given this dependence, the system cannot be easily applied to reports written in other languages. And even if the system was not dependable on an English lexicon, it is not certain that the results would be the same if another language was used, because of, for example, differences in syntax (other examples of tools developed focused on English include (5), (6) and (7)). This have been an obstacle in the sharing of Radiology information between different communities, which is important to understand and effectively address health problems.

There are mainly two possible solutions to this problem. One is to translate the lexicon itself (8,9) and the other is to translate the reports. MRA (Multilingual Radiology Annotator) is a web application that explores the implementation of this last solution. It translates and annotates Radiology text with RadLex terms, a Named-Entity Recognition (NER) task. This is a relevant task since the outputs from NER systems can be used in Image Retrieval (10) and Information Retrieval (11) systems and can be useful for improving automatic Question Answering (12). Basically, MRA is a prototype of what can be done when integrating translation in medical applications.

# Description

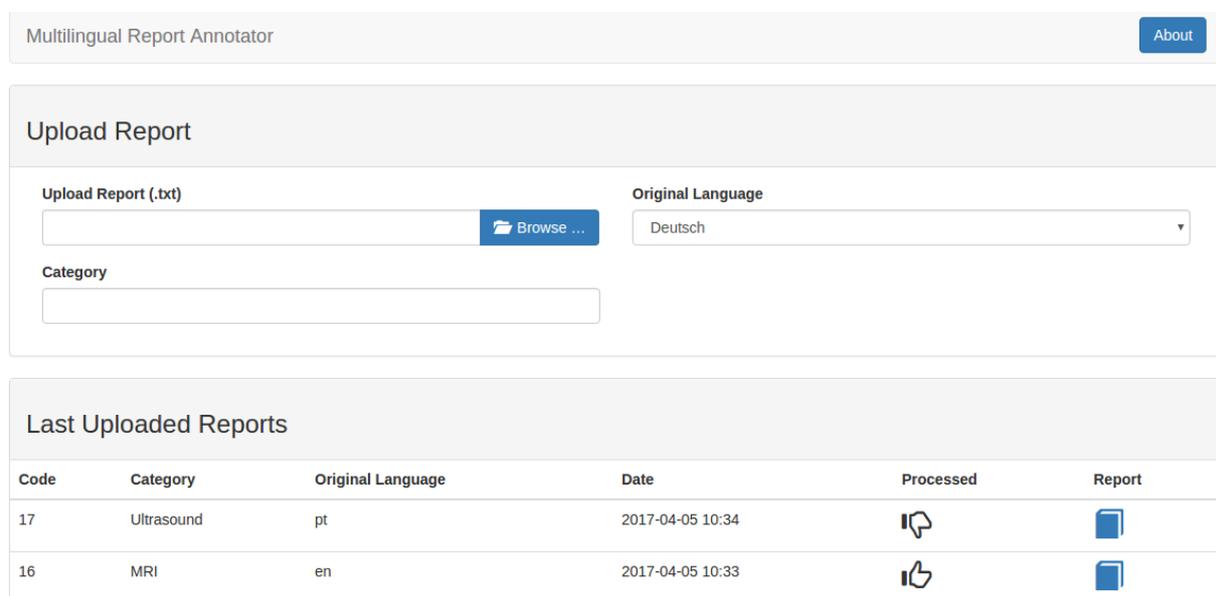

*Figure 1: MRA Homepage*

The flow of the application goes like this. The user uploads a text file containing Radiology text to the application (Figure 1), and, if the text is in a non-English language, it is sent to Unbabel's[1] translation services, which combines human translation with machine translation. The application currently supports five languages beyond English, the ones supported by Unbabel API. In this prototype only machine translation is being used, for demonstration purposes. In a real-life scenario, human translation could also be used for more reliable results. So, the text is sent to translation and after a while (approximately 2 minutes, to simulate a real human and machine translation) the translation is ready. Then, the translated text is sent to BioPortal[2] annotation services. After this is done it is possible to explore the annotations in the translated text (Figure 2). The interface of the annotations was partly inspired by a similar project called LexMap (13).

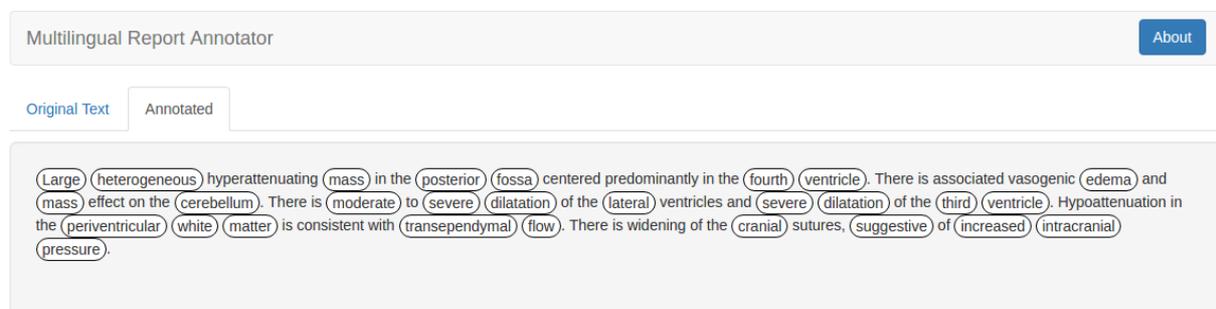

Figure 2: MRA Annotations Interface

The software, as it is, its dependable on the user having an Unbabel's and a BioPortal's API keys. The BioPortal API key is easy to obtain. If you do not have and cannot obtain an Unbabel API key, it will not be hard do adapt the software so that you can use another translation service.

The back-end was developed using Python's Flask[3] web-framework. It makes use of Celery[4] to handle the requests for translations and annotations. The software code and a guide for its installation can be found at the following GiHub repository: https://github.com/lasigeBioTM/MRA.

---

1 https://unbabel.com/
2 http://bioportal.bioontology.org/annotator
3 http://flask.pocoo.org/
4 http://www.celeryproject.org/

# Conclusion

The idea is that this application can be used to bootstrap other, more useful applications. It can also be shown to clinical institutions to pique their interest on what can be done when you integrate translation in medical practice.

# Acknowledgements

Thanks to Miguel Rodrigues and André Lamúrias for suggestions.

# References

1.  Pons, E., Braun, L. M. M., Hunink, M. G. M., et al. (2016) Natural Language Processing in Radiology: A Systematic Review, *Radiology*, **279**, 329–343.

2.  Meystre, S. M., Savova, G. K. and Hurdle, J. F. (2008) Extracting Information from Textual Documents in the Electronic Health Record : A Review of Recent Research, *IMIA Yearb. Med. Informatics*, **35**, 128–144.

3.  Hassanpour, S. and Langlotz, C. P. (2016) Information extraction from multi-institutional radiology reports, *Artif. Intell. Med.*, **66**, 29–39.

4.  Langlotz, C. P. (2006) RadLex: a new method for indexing online educational materials, *RadioGraphics*, **26**, 1595–1597.

5.  Ananda-Rajah, M. R., Martinez, D., Slavin, M. A., et al. (2014) Facilitating surveillance of pulmonary invasive mold diseases in patients with haematological malignancies by screening computed tomography reports using natural language processing, *PLoS One*, **9**, 1–8.

6.  Martinez, D., Ananda-Rajah, M. R., Suominen, H., et al. (2015) Automatic detection of patients with invasive fungal disease from free-text computed tomography (CT) scans, *J. Biomed. Inform.*, **53**, 251–260.

7.  Yetisgen-Yildiz, M., Gunn, M. L., Xia, F., et al. (2013) A text processing pipeline to extract recommendations from radiology reports, *J. Biomed. Inform.*, **46**, 354–362.

8.  Bretschneider, C., Oberkampf, H., Zillner, S., et al. In *Proceedings of the Third Workshop on Semantic Web and Information Extraction*; 2014; pp. 1–8.

9.  Cotik, V., Filippo, D. and Castaño, J. (2015) An Approach for Automatic Classification of Radiology Reports in Spanish, *Stud. Health Technol. Inform.*, **216**, 634–638.

10. Gerstmair, A., Daumke, P., Simon, K., et al. (2012) Intelligent image

retrieval based on radiology reports, *Eur. Radiol.*, **22**, 2750–2758.

11. Antony, B. J. and Suryanarayanan Mahalakshmi, G. (2015) Content-based Information Retrieval by Named Entity Recognition and Verb Semantic Role Labelling, *J. Univers. Comput. Sci.*, **21**, 1830–1848.

12. Toral, A., Noguera, E., Llopis, F., et al. In *Natural Language Processing and Information Systems, Proceedings*; Springer, Berlin, Heidelberg, 2005; Vol. 3513, pp. 181–191.

13. Hostetter, J., Wang, K., Siegel, E., et al. (2015) Using Standardized Lexicons for Report Template Validation with LexMap, a Web-based Application, *J. Digit. Imaging*, **28**, 309–314.